\documentclass{article}

\usepackage[final,nonatbib]{neurips_2024}

\usepackage[utf8]{inputenc} % allow utf-8 input
\usepackage[T1]{fontenc}    % use 8-bit T1 fonts
\usepackage{hyperref}       % hyperlinks
\usepackage{url}            % simple URL typesetting
\usepackage{booktabs}       % professional-quality tables
\usepackage{amsfonts}       % blackboard math symbols
\usepackage{nicefrac}       % compact symbols for 1/2, etc.
\usepackage{microtype}      % microtypography
\usepackage{xcolor}         % colors

\usepackage{amsmath}
\usepackage{graphicx}
\usepackage{float}
\usepackage{hhline}
\usepackage{verbatim}
\usepackage[T1]{fontenc}
\usepackage{multirow}
\usepackage{tabularx}
\usepackage{makecell}
\usepackage{marvosym}
\usepackage{arydshln}

\title{Surgical SAM 2: Real-time Segment Anything in Surgical Video by Efficient Frame Pruning}

\author{%
    Haofeng Liu$^{1}$\thanks{Equal contributions}~~, 
    Erli Zhang$^{1 \star}$, 
    Junde Wu$^{2 \star}$, 
    Mingxuan Hong$^{1}$, 
    Yueming Jin$^1$\thanks{Corresponding author (\texttt{ymjin@nus.edu.sg}) }\\
    $^1$National University of Singapore\\
    $^2$University of Oxford\\
    \texttt{haofeng.liu@u.nus.edu}, \texttt{erli.zhang@u.nus.edu}, \texttt{jundewu@ieee.org}\\ 
    \texttt{mingxuan.hong@u.nus.edu}, \texttt{ymjin@nus.edu.sg}
}
\begin{document}

\maketitle

\begin{abstract}
  Surgical video segmentation is a critical task in computer-assisted surgery and is vital for enhancing surgical quality and patient outcomes. Recently, the Segment Anything Model 2 (SAM2) framework has shown superior advancements in image and video segmentation. However, SAM2 struggles with efficiency due to the high computational demands of processing high-resolution images and complex and long-range temporal dynamics in surgical videos. To address these challenges, we introduce Surgical SAM 2 (SurgSAM2), an advanced model to utilize SAM2 with an Efficient Frame Pruning (EFP) mechanism, to facilitate real-time surgical video segmentation. The EFP mechanism dynamically manages the memory bank by selectively retaining only the most informative frames, reducing memory usage and computational cost while maintaining high segmentation accuracy. Our extensive experiments demonstrate that SurgSAM2 significantly improves both efficiency and segmentation accuracy compared to the vanilla SAM2. Remarkably, SurgSAM2 achieves a 3$\times$ FPS compared with SAM2, while also delivering state-of-the-art performance after fine-tuning with lower-resolution data. These advancements establish SurgSAM2 as a leading model for surgical video analysis, making real-time surgical video segmentation in resource-constrained environments a reality. Our source code is available at https://github.com/jinlab-imvr/Surgical-SAM-2.
\end{abstract}

\section{Introduction}
Surgical video scene segmentation is a critical task in computer-assisted surgery, where the precise identification and delineation of surgical instruments and tissues within video sequences are essential. This capability underpins various applications, such as instrument tracking and pose estimation, intraoperative guidance, and postoperative analysis~\cite{ref_robu2020tow,ref_du2019pba}, ultimately enhancing surgical precision, reducing operative times, and improving patient outcomes. Achieving high segmentation accuracy with low computational cost is vital in real-world deployment considering limited resources in clinical centers. Real-time prediction also plays a core role in practical applications, to enable the method to provide timely decision-making support and navigation, generate real-time warning of potential deviations and anomalies, and facilitate remote surgical proctoring and team communication~\cite{ref_gonzalez2020in}. However, accurate and efficient surgical video segmentation is challenging, due to the highly complicated surgical scene with lighting reflection and occlusion from blood and smoke, limited inter-class variance with highly similar appearance (e.g., different instruments), long-range duration with dynamic contexts, irregular artifacts, and noises.

The Segment Anything Model 2 \cite{ref_Yu2024SAM2}, with its ViT architecture and multi-scale feature extraction, has significantly advanced image and video segmentation, and, with meaningful modifications, proves to be highly effective for both 2D and 3D medical image segmentation~\cite{medsam2}. However, its application to surgical videos presents challenges, particularly in terms of computational efficiency. Surgical videos, which are typically high-resolution and last over several hours, require the effective modeling of long-range spatiotemporal dynamics. Despite its strengths, SAM2’s substantial computational demands undermine its suitability for such scenarios. This challenge is further exacerbated in surgical settings, where the redundancy of semantic information across frames is prevalent. Especially, considering that the camera generally remains fixed or moves slowly during an operation, the dominant tissues show high similarity across frames, leading to unnecessary processing of repetitive data~\cite{ref_jin2020}. Additionally, SAM2’s memory bank mechanism, which stores frames sequentially, frequently retains redundant information, further inflating computational costs and hindering real-time application in surgery.

In the context of real-time surgical video segmentation, efficiency is not merely desirable but a critical necessity. Current approaches, including SAM2, have not adequately addressed this need, particularly in resource-constrained surgical environments where computational resources are limited. To address these challenges, we introduce Surgical SAM 2 (SurgSAM2), one of the first works to specifically optimize SAM2 for surgical video segmentation, improving efficiency without compromising accuracy.

SurgSAM2 is optimized for real-time application by integrating a dynamic memory bank management mechanism that reduces redundancy and computational load. This mechanism intelligently controls the retention of video frames by selectively pruning those that contribute less to ongoing analysis, thereby minimizing the impact of redundant frames that might otherwise dilute attention on critical objects. Unlike the original SAM2, which retains frames based on their order of arrival, SurgSAM2 employs a dynamic scoring mechanism based on cosine similarity to assess the importance of each frame. By strategically reducing the memory bank size, SurgSAM2 achieves state-of-the-art results, significantly increasing prediction speed in frames per second (FPS) and reducing memory requirements. This optimization ensures that SurgSAM2 provides reliable and efficient performance in complex and dynamic scenario of surgical video analysis, making it a leading solution for real-time surgical video segmentation.

We conducted extensive experiments on instrument segmentation of surgical video, with two widely recognized benchmarks: the EndoVis17 and EndoVis18 datasets. The results demonstrate that SurgSAM2 maintains high performance despite resource constraints, making it an ideal solution for real-time surgical video segmentation. This efficiency not only enables more accurate and timely decision-making during surgeries but also has the potential to reduce operating times and improve patient outcomes by ensuring that critical information is processed swiftly and accurately. Moreover, while our primary focus is on surgical video segmentation, the principles behind SurgSAM2's design have the potential to be adapted to broader video analysis tasks.

In summary, our contributions are as follows:
\begin{enumerate}
\item We introduce SurgSAM2, a pioneering SAM2-based work for surgical video segmentation, designed to meet the specific needs of the surgical domain.
\item We propose an innovative approach to optimize SAM2 for surgical environments by integrating a dynamic memory management module that leverages selective frame pruning to reduce computational demands while maintaining high accuracy.
\item We extensively evaluate SurgSAM2 on the EndoVis17 and EndoVis18, demonstrating its superior performance in both accuracy and efficiency of surgical instrument segmentation.
\end{enumerate}

\section{Related work}

\subsection{Surgical Instrument Segmentation}
The field of surgical instrument segmentation has evolved significantly with deep learning, particularly through fully convolutional networks (FCNs) and encoder-decoder architectures like U-Net, which laid the foundation for the domain~\cite{ref_shvets2018ternausnet}. However, these early methods often faced challenges in dynamic surgical environments, struggling with spatial inconsistencies and the complex interactions between instruments and surrounding anatomy~\cite{ref_ni2020feature,ref_jin2019incorporating,ou2022mvd}.

To address these challenges, recent advancements have focused on transformer-based models and attention mechanisms, such as Swin Transformers and multi-scale attention U-Nets, which offer better robustness and adaptability in handling the complex visual features of surgical instruments~\cite{ref_swin2021,ref_multiscale2020}. Following the introduction of the Segment Anything Model (SAM)\cite{ref_kirillov2023sam} and SAM2\cite{ref_Yu2024SAM2}, there has been a shift towards models tailored specifically for the medical image segmentation~\cite{msa} and surgical video segmentation, such as SurgicalSAM~\cite{ref_Yue2024surgsam}. Despite these advances, challenges remain, particularly in achieving efficient processing within resource constraints typical of surgical settings. This gap in efficiency is the core motivation behind SurgSAM2, which seeks to optimize performance for real-time surgical applications~\cite{ref_marie2021surgical,ref_van2021deep}.

\subsection{Segment Anything Model 2}
SAM2 builds on Vision Transformers (ViTs) with enhanced multi-scale feature extraction, making it a powerful tool in image and video segmentation~\cite{ref_Yu2024SAM2}. Following targeted modifications, SAM2 also shows substantial effectiveness in the 2D and 3D medical image segmentation~\cite{medsam2}. However, its use in surgical video segmentation faces significant challenges due to the computational intensity of ViTs, which require substantial resources, limiting their practicality in real-time, resource-constrained environments~\cite{ref_fastsam}. SAM2’s reliance on a first-come-first-serve memory mechanism exacerbates inefficiencies, as it retains potentially redundant frames, further slowing down processing. The need for optimized models that reduce computational overhead while maintaining strong segmentation performance is critical, paving the way for more efficient solutions like SurgSAM2.

\subsection{Memory Bank Restriction}
Efficient memory management is crucial for real-time applications, especially in the context of surgical video segmentation where computational resources are limited. Strategies like XMem~\cite{ref_XMem} and RMem~\cite{ref_RMem} have explored ways to prioritize and retain only the most relevant frames during video analysis. Building on these approaches, SurgSAM2 introduces an efficient frame pruning mechanism, which uses a cosine similarity-based scoring system to retain only the most informative frames, reducing memory usage and increasing processing speed. This approach directly addresses inefficiencies in SAM2 memory management, making SurgSAM2 better suited to the fast-paced and resource-constrained demands of real-time surgical video analysis.

\section{Methods}

SurgSAM2 is an advanced model designed specifically for the complex and resource-constrained environments of surgical video segmentation. Building upon the SAM2~\cite{ref_Yu2024SAM2}, SurgSAM2 incorporates a dynamic memory bank management mechanism to optimize the retention and use of video frames during segmentation tasks. This innovation not only reduces the computational load but also enhances segmentation accuracy by selectively retaining the most relevant and useful information. The memory bank consists of the current frame and dynamically selected preceding frames, which are critical for maintaining temporal context. By integrating these advancements, SurgSAM2 addresses the unique challenges posed by real-time surgical video analysis, offering a robust solution that balances efficiency and performance.

\begin{figure}
    \centering
    \includegraphics[width=0.99\linewidth]{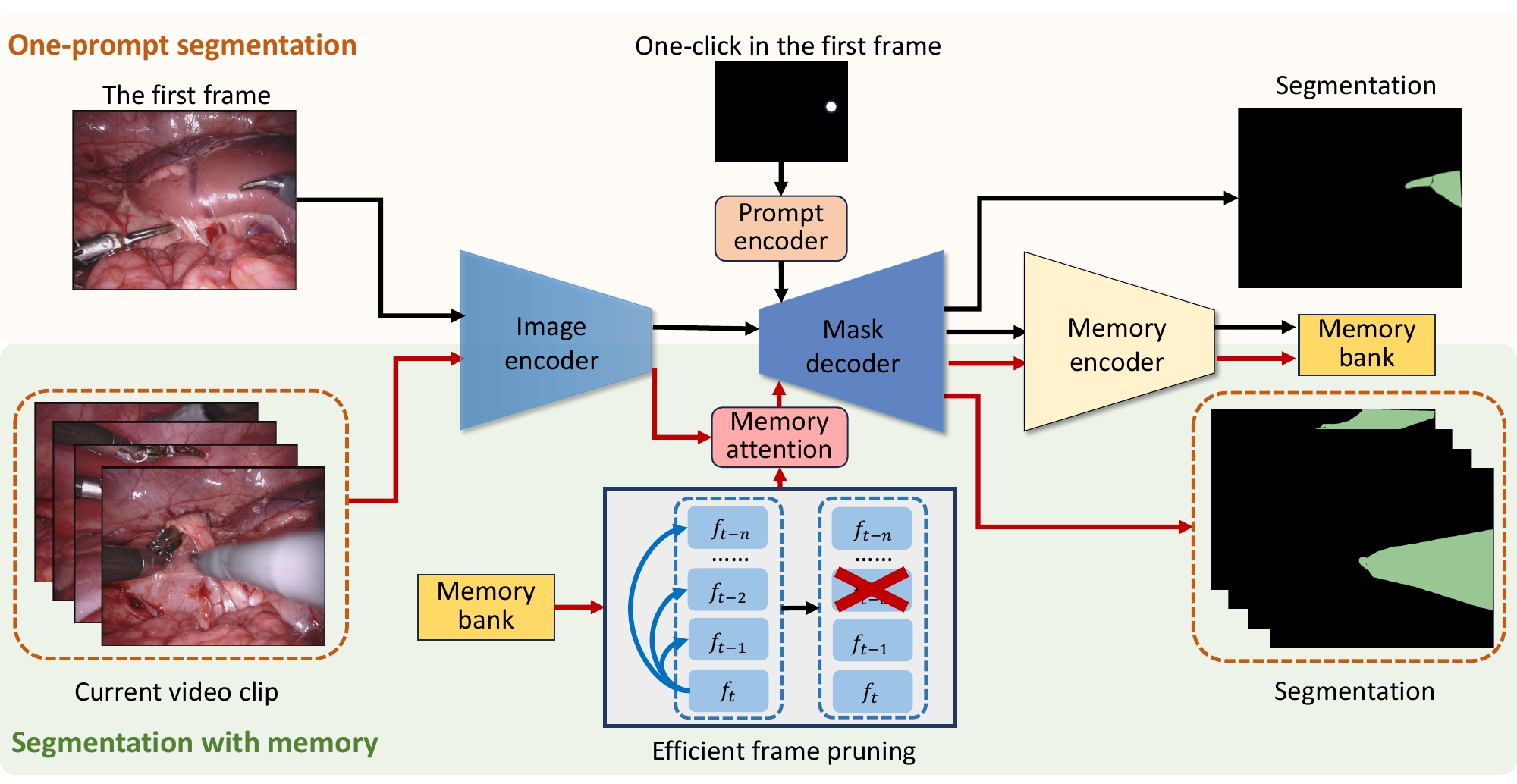}
    \vspace{-10pt}
    \caption{Architecture of the proposed model SurgSAM2.}
    \label{fig:architecture}
    \vspace{-15pt}
\end{figure}

\subsection{SurgSAM2 Architecture}

SurgSAM2 builds upon the SAM2~\cite{ref_Yu2024SAM2}, leveraging its robust ViT architecture specifically adapted for surgical video segmentation. The backbone of SAM2 is retained in SurgSAM2 but with significant optimizations to meet the unique challenges posed by surgical environments.

At its core, the image encoder of SurgSAM2 processes incoming video frames into detailed embeddings, capturing both local and global features necessary for accurate segmentation. While this architecture is consistent with the vanilla SAM2, SurgSAM2 enhances it by integrating a dynamic memory management system that selectively prunes less relevant frames, ensuring that only the most critical data is retained for further analysis.

These refinements allow SurgSAM2 to maintain the high performance associated with SAM2 while significantly improving efficiency, making it better suited for real-time applications in resource-constrained surgical settings.

\subsection{Efficient Frame Pruning}

A key innovation in SurgSAM2 is the implementation of an efficient frame pruning mechanism, designed to intelligently manage which video frames are retained for further processing. This mechanism dynamically evaluates the relevance of each incoming frame before it is added to the memory bank, ensuring that only the most informative frames are preserved while discarding those that contribute minimally to the segmentation task.

For each frame \( f_t \) that is passed into the memory bank, the cosine similarity \( S(f_t, f_i) \) between \( f_t \) and the past \( n \) frames \( \{f_{t-n}, \dots, f_{t-2}, f_{t-1}\} \) in the sliding window is computed as follows:

\begin{equation}
S(f_t, f_i) = \frac{f_t \cdot f_i}{\|f_t\| \|f_i\|}
\end{equation}

After computing the \( n \) cosine similarities, the mechanism identifies the \( m \) most similar frames from these \( n \) frames. These \( m \) frames are pruned, and the remaining $(n-m)$ frames plus frame \( f_t \) are then stored in the memory bank for memory cross-attention. Notably, the first frame \( f_0 \), which serves as a key reference, is always retained in the memory bank and is not counted towards the dynamic memory bank size. This ensures that the memory bank always includes a critical reference frame while still optimizing for efficiency by retaining only the most relevant subsequent frames.

Considering that the vanilla SAM2 model uses a memory bank size of six past frames and the first reference frame for spatiotemporal modeling, our configuration is set as \( n = 5 \) and \( m = 2 \). 
We first compute the cosine similarity between frame \( f_t \) and the past \( n \) frames. 
Subsequently, we prune the two most similar frames \( m = 2 \), leaving four frames in the dynamic memory bank, which aligns with our configuration for the EFP mechanism in SurgSAM2.

By implementing this selective EFP strategy, SurgSAM2 effectively reduces memory usage and computational load, allowing the model to process video frames more efficiently. In surgical videos, where scenes often exhibit high visual similarity and repeated semantic content across frames, this approach is particularly effective in eliminating redundancy while maintaining efficiency. Despite the reduction in stored frames, the model maintains high segmentation accuracy by focusing computational resources on the most critical data, making it well-suited for real-time surgical video analysis in resource-constrained environments.

\subsection{Implementation Details}
All experiments were conducted on an RTX A6000 GPU with 48GB using the ViT-Small backbone. The precision for the experiments was set to \texttt{bfloat16}, which helps to reduce computational load while maintaining model performance. Note that the vanilla SAM2 uses a resolution of $1024\times1024$ and ViT-Base+. However, video training at this resolution is highly time-consuming and difficult to stabilize with limited computational resources in the surgical scenario. Therefore, we opted to fine-tune using a resolution of $512\times512$ and adopted the weights of the ViT-Small version from SAM2.
Following the training strategy outlined in SAM2, we alternated between video and image training. To fully explore the potential of SAM2, we train the multi-mask output, iou prediction, and occlusion prediction. To preserve the generalization ability of SAM2, we fine-tuned only the mask decoder and memory module, keeping the prompt encoder and image encoder frozen. 

For video training, a batch size of 12 was used, with each batch containing 8 frames and up to 3 objects per frame. In image training, the batch size was set to 32, with a maximum of 3 objects per image. This alternation between video and image data during training ensures that the model effectively learns from both dynamic and static content, enhancing its generalization capabilities.
We utilized a learning rate of \(2 \times 10^{-4}\) for the mask decoder and \(2 \times 10^{-5}\) for the memory encoder. As for the video data augmentation strategy, we followed the strategy of Cutie~\cite{ref_cutie}, which is designed to improve the model robustness by simulating various challenging scenarios during training. During inference, we re-scaled the output segmentation to its original resolution for fair evaluation. To ensure high-quality segmentation for the first frame—crucial for effective instrument tracking in subsequent frames—we employed the original resolution (1024) for the first frame, followed by lower resolution (512) processing for the remaining frames to optimize model efficiency.

\section{Experiment}

\subsection{Dataset}

We extensively evaluate the proposed SurgSAM2 model on two widely recognized and publicly available datasets: the 2017 MICCAI EndoVis Instrument Challenge (EndoVis17)~\cite{ref_Allan2019EndoVis17} and the 2018 MICCAI EndoVis Scene Segmentation Challenge (EndoVis18)~\cite{ref_Allan2020EndoVis18}. EndoVis18 presents more complicated surgical scenes, therefore proving more challenging compared to EndoVis17. The EndoVis17 dataset comprises eight training videos each containing 225 frames, eight testing videos collected followed by training videos, and another two hold-out testing videos (sequence 9, 10), comprising 1200 frames in total. The EndoVis18 dataset consists of 15 videos with each consisting of 149 frames. For Endovis17, we use the hold-out test set for evaluation. For EndoVis18, we split the sequences of 2, 5, 9, and 15 for testing following the standard procedure in ISINet~\cite{ref_gonzalez2020in}.

The data from the EndoVis17 and EndoVis18 datasets were pre-processed following the approach described by Shvets et al.~\cite{ref_shvets2018ternausnet}. Given that the EndoVis17 and EndoVis18 datasets from ISINet~\cite{ref_gonzalez2020in} only include instrument-type labels rather than instance-level labels, we re-annotate the data to ensure that our model is evaluated on a more detailed and instance-specific level, allowing for precise instrument segmentation and better generalization across different surgical scenarios.

\subsection{Evaluation Metrics}

To comprehensively evaluate the performance of SurgSAM2, we adopt numerous and widely-used evaluation metrics in video object segmentation (VOS) that assess both the accuracy and computational efficiency of segmentation~\cite{Perazzi2016}. These metrics are chosen to provide a holistic view of the model’s capabilities in the context of surgical video segmentation. For these metrics, we follow the evaluation protocol in the video object segmentation benchmark, specifically excluding the first and last frames from the assessment. 
We also utilize the official evaluation protocol in EndoVis Challenge~\cite{ref_Allan2019EndoVis17,ref_Allan2020EndoVis18} for method validation.

\noindent\textbf{Intersection-over-Union (J or IoU):} Intersection-over-Union (IoU), denoted as \(J\), measures the overlap between the predicted segmentation and the ground truth. It is calculated as the ratio of the intersection between the predicted and true positive regions to their union. IoU is a standard metric in segmentation tasks, offering a robust measure of the accuracy of the model's predictions.

\noindent\textbf{Boundary F1 Score (F):} The Boundary F1 Score, denoted as \(F\), assesses the accuracy of the predicted boundaries in the segmentation mask. It calculates the F1 score specifically along the edges of the segmented regions, providing insight into how well the model captures the precise contours of the surgical instruments.

\noindent\textbf{J\&F Score:} The J\&F score is a composite metric that averages the IoU and Boundary F1 scores, providing a balanced assessment of both region overlap and boundary accuracy. This metric is particularly useful for evaluating segmentation quality in tasks where both precise region delineation and boundary accuracy are crucial.

\noindent\textbf{Dice Coefficient (Dice):} The Dice Coefficient, another widely used metric in segmentation tasks, measures the similarity between the predicted segmentation and the ground truth. It is closely related to IoU but places more emphasis on the overlap, making it a complementary metric to IoU.

\noindent\textbf{Challenge IoU (CIoU):} The Challenge IoU metric follows the evaluation protocol outlined in the EndoVis18 Challenge~\cite{ref_Allan2020EndoVis18}. CIoU calculates the Intersection over Union (IoU) for each frame individually, considering only the objects present in that specific frame. The IoU scores are then averaged across all frames to produce the final CIoU score, providing a more precise assessment of segmentation performance in the context of dynamic and frame-specific object presence.

\noindent\textbf{Frames Per Second (FPS):} To evaluate the real-time performance of SurgSAM2, we measure the FPS during inference. This metric is crucial for applications in surgical environments where timely processing of video frames is essential for effective decision-making.

\noindent\textbf{Memory Usage:} We also assess the memory efficiency of SurgSAM2 by calculating the model's memory footprint during inference. Given the resource-constrained nature of many surgical settings, reducing memory usage is a key objective of our approach. By optimizing the memory bank size and employing selective frame retention, SurgSAM2 achieves a balance between high segmentation accuracy and efficient memory usage.

These metrics together provide a comprehensive evaluation of SurgSAM2, highlighting its strengths in segmentation accuracy, boundary precision, computational speed, and memory efficiency. By balancing these aspects, SurgSAM2 is well-positioned to meet the demands of real-time surgical video analysis.

\subsection{Experimental Results}

\subsubsection{Evaluation on Model Efficiency}

We conducted a thorough evaluation of SurgSAM2's performance in terms of FPS and memory usage, comparing it with the vanilla SAM2 across various configurations. The results, as illustrated in Table~\ref{tab:full_mask_comparison}, ~\ref{tab:one_point_comparison} and~\ref{tab:five_points_comparison}, clearly demonstrate that by implementing a cosine similarity-based efficient frame pruning mechanism and reducing the memory bank size, both FPS and memory efficiency are significantly improved.

\begin{table*}[!ht]
\centering
% \scriptsize
\small
\vspace{-10pt}
\caption{Performance comparison on EndoVis17 and EndoVis18 datasets for Full Mask setting.}
\vspace{5pt}

\renewcommand{\arraystretch}{1.2}
\setlength{\tabcolsep}{4pt} % Adjust the horizontal space between columns
\begin{tabular}{|c|c|c|c|c|c|c|c|c|c|}
\hline
\textbf{Dataset} & 
\textbf{Method} &
\textbf{EFP} &
\textbf{\makecell{Fine-\\tuning}} &
\textbf{J} & \textbf{F} & \textbf{J\&F} & \textbf{Dice} & \textbf{FPS} & \textbf{\makecell{Memory \\ (GB)}} \\ \hline
\multirow{3}{*}{\makecell{Endovis\\17}} & SAM2 & No & No & 85.9 & 89.1 & 87.5 & 90.2 & 29.10 & 3.10 \\ \cdashline{2-10}
& Ours & Yes & No & 85.7 & 88.6 & 87.2 & 89.9 & 33.00 & 2.83 \\ \cdashline{2-10}
& Ours & Yes & Yes & \textbf{88.2} & \textbf{90.6} & \textbf{89.4} & \textbf{92.3} & \textbf{86.03} & \textbf{1.08} \\ \hline
\multirow{3}{*}{\makecell{Endovis\\18}} & SAM2 & No & No & 78.4 & 78.6 & 78.5 & 81.7 & 29.18 & 3.14 \\ \cdashline{2-10}
& Ours & Yes & No & 81.9 & 81.9 & 81.9 & 85.2 & 33.08 & 2.82 \\ \cdashline{2-10}
& Ours & Yes & Yes & \textbf{81.9} & \textbf{82.0} & \textbf{82.0} & \textbf{85.3} & \textbf{86.11} & \textbf{1.02} \\ \hline
\end{tabular}
% % \vspace{5pt}

% \vspace{-15pt}
\label{tab:full_mask_comparison}
\end{table*}

\begin{table*}[!ht]
\centering
% \scriptsize
\small
% \footnotesize
\vspace{-10pt}
\caption{Performance comparison on EndoVis17 and EndoVis18 datasets for One Point setting.}
\vspace{5pt}
% \vspace{-25pt}
\renewcommand{\arraystretch}{1.2}
\setlength{\tabcolsep}{4pt} % Adjust the horizontal space between columns
\begin{tabular}{|c|c|c|c|c|c|c|c|c|c|}
\hline
\textbf{Dataset} & 
\textbf{Method} &
\textbf{EFP} &
\textbf{\makecell{Fine-\\tuning}} &
\textbf{J} & \textbf{F} & \textbf{J\&F} & \textbf{Dice} & \textbf{FPS} & \textbf{\makecell{Memory \\ (GB)}} \\ \hline

\multirow{3}{*}{\makecell{Endovis\\17}} 
& SAM2 & No & No & 81.1 & 83.8 & 82.5 & 85.1 & 29.09 & 3.11 \\ \cdashline{2-10}
& Ours& Yes & No & 79.9 & 83.3 & 81.6 & 84.8 & 33.16 & 2.89 \\ \cdashline{2-10}
& Ours& Yes & Yes & \textbf{82.7} & \textbf{84.2} & \textbf{83.4} & \textbf{87.3} & \textbf{85.95} & \textbf{1.09} \\ \hline

\multirow{3}{*}{\makecell{Endovis\\18}} 
& SAM2 & No & No & 71.5 & 73 & 72.3 & 74.2 & 29.21 & 3.15 \\ \cdashline{2-10}
& Ours & Yes & No & 71.9 & 73.3 & 72.6 & 75.5 & 33.07 & 2.85 \\ \cdashline{2-10}
& Ours & Yes & Yes & \textbf{72.6} & \textbf{73.8} & \textbf{73.2} & \textbf{76.7} & \textbf{86.04} & \textbf{1.04} \\ \hline
\end{tabular}
% \vspace{5pt}
% \vspace{-15pt}
\label{tab:one_point_comparison}
\end{table*}

\begin{table*}[!ht]
\centering
% \scriptsize
\small
\vspace{-10pt}
\caption{Performance comparison on EndoVis17 and EndoVis18 datasets for Five Points setting.}
\vspace{5pt}
% \vspace{-25pt}
\renewcommand{\arraystretch}{1.2}
\setlength{\tabcolsep}{4pt} % Adjust the horizontal space between columns
\begin{tabular}{|c|c|c|c|c|c|c|c|c|c|}
\hline
\textbf{Dataset} & 
\textbf{Method} &
\textbf{EFP} &
\textbf{\makecell{Fine-\\tuning}} &
\textbf{J} & \textbf{F} & \textbf{J\&F} & \textbf{Dice} & \textbf{FPS} & \textbf{\makecell{Memory \\ (GB)}} \\ \hline

\multirow{3}{*}{\makecell{Endovis\\17}} 
& SAM2 & No & No & 82.0 & 85.9 & 83.9 & 86.9 & 29.13 & 3.14 \\ \cdashline{2-10}
& Ours & Yes & No & 81.9 & 85.3 & 83.6 & 86.7 & 33.08 & 2.85 \\ \cdashline{2-10}
& Ours & Yes & Yes & \textbf{86.9} & \textbf{89.1} & \textbf{88.0} & \textbf{91.4} & \textbf{85.94} & \textbf{1.05} \\ \hline

\multirow{3}{*}{\makecell{Endovis\\18}} 
& SAM2 & No & No & 76.2 & 76.3 & 76.3 & 80 & 29.29 & 3.12 \\ \cdashline{2-10}
& Ours & Yes & No & 79.1 & 79.2 & 79.1 & 82.8 & 33.13 & 2.87 \\ \cdashline{2-10}
& Ours & Yes & Yes & \textbf{80.9} & \textbf{80.7} & \textbf{80.8} & \textbf{84.9} & \textbf{86.00} & \textbf{1.02} \\ \hline
\end{tabular}
% \vspace{5pt}
\vspace{-15pt}
\label{tab:five_points_comparison}
\end{table*}

\begin{comment}

\end{comment}

In evaluating the efficiency of SurgSAM2 compared to the vanilla SAM2, we observed consistent improvements across different prompt settings on both the EndoVis17~\cite{ref_Allan2019EndoVis17} and EndoVis18~\cite{ref_Allan2020EndoVis18} datasets. On average, SurgSAM2 demonstrated a 13.8\% increase in FPS and an 8.5\% reduction in memory usage across the various prompt settings (Full Mask, One Point, and Five Points). These results underscore the effectiveness of our cosine similarity-based frame pruning mechanism in enhancing computational efficiency, particularly in resource-constrained environments. 

\subsubsection{Evaluation on Model Accuracy}

Apart from model efficiency, we also evaluated SurgSAM2's performance compared with the vanilla SAM2 model. The results, as detailed in Table~\ref{tab:full_mask_comparison}, ~\ref{tab:one_point_comparison} and~\ref{tab:five_points_comparison}, indicate that SurgSAM2 consistently outperforms the vanilla SAM2 across various settings and datasets.

\begin{comment}

\end{comment}

From the experiment results, it is shown that reducing the memory bank size while applying the cosine similarity mechanism led to mixed results in segmentation accuracy for the EndoVis17 dataset. Specifically, while the J\&F metric saw a slight decrease of 0.5\% and the Dice score dropped by 0.3\% compared to the original memory bank configuration, there was an improvement in FPS and memory efficiency. On the other hand, for the more challenging dataset EndoVis18, SurgSAM2 achieved a 2.2\% increase in the J\&F metric and a 2.5\% improvement in the Dice score, reflecting the positive impact of the proposed EFP mechanism.

This outcome can be explained by the principles of memory management in video segmentation models. In larger memory banks, the accumulation of redundant information can dilute the attention scores between relevant objects, leading to less precise segmentation. By reducing the memory bank size, the model becomes more selective, retaining only the most informative frames. This selective retention helps maintain higher attention scores between correct objects, effectively pruning redundant frames. Consequently, while there may be a slight reduction in segmentation accuracy, the model gains significant efficiency, which is crucial for real-time applications.

However, it is important to note that while a smaller memory bank can improve performance by reducing noise, it should be large enough to store all useful information. Striking the right balance is crucial for maintaining a precise and efficient representation of temporal information, ultimately improving overall performance in surgical video segmentation. 
The efficiency gains observed with SurgSAM2 underscore the significance of this balance, as the model achieves improved FPS and reduced memory usage without substantially compromising segmentation accuracy.

\subsubsection{Fine-Tuning for Optimized Segmentation and Efficiency}
We further investigated the efficacy of fine-tuning in our SurgSAM2 model. Typically, higher input resolutions enhance segmentation accuracy. 
Surprisingly, we found SurgSAM2, which predicts segmentation based on half the original resolution, (512, in our setting), can already outperform the vanilla SAM2 at the full resolution. This result is important in the surgical scenario because a smaller resolution allows for real-world model training with reduced memory requirements. Most importantly, this enables a dramatic increase in prediction speed for real-time segmentation of surgical video. In the 1-point setting, which requires very little effort from the surgeon—just a single click on the first frame of the entire surgical procedure—our SurgSAM2 can increase precision from 85.1\% Dice to 87.3\%, and also efficiency from 29 to 86 FPS in EndoVis17, as in shown in Table~\ref{tab:one_point_comparison}.

Looking at all three tables~\ref{tab:full_mask_comparison}, ~\ref{tab:one_point_comparison} and~\ref{tab:five_points_comparison} into details, we can see that our SurgSAM2 with fine-tuning shows a marked and consistent improvement in segmentation accuracy across different settings. For instance, for Five points, our SurgSAM2 with fine-tuning led to an increase in the J\&F metric from 83.6\% to 88.0\% and the Dice score from 86.9\% to 91.4\% in the EndoVis17 dataset. Similarly, our method resulted in an increase in the J\&F metric from 76.3\% to 80.8\% and the Dice score from 80\% to 84.9\% in the EndoVis18 dataset. This significant enhancement underscores the effectiveness of our fine-tuning strategy in conditions that closely mimic clinical practice, further validating SurgSAM2's capability to deliver precise and consistent segmentation in practical medical applications.

\subsubsection{Comparative Model Evaluation}
We compared SurgSAM2 against the state-of-the-art methods specifically designed to surgical instrument segmentation and some advanced SAM-based methods, utilizing the Challenge IoU metric from EndoVis18 to assess performance. Results of other methods are quoted from their papers~\cite{ref_Yue2024surgsam}. Note that a completely fair comparison cannot be achieved, as most of these existing methods do not require prompts in inference. Additionally, most methods for instrument segmentation are designed for type segmentation, which does not need to distinguish different instances of the same type, though this is one of the most challenging problems in surgical instrument segmentation.
Our method aims at a more practical setting to segment instruments on an instance level. We also compared the vanilla SAM2 in this setting and listed all the results for the EndoVis18 dataset in Table~\ref{tab:iou_comparison}.

\begin{table*}[!ht]
% \scriptsize

\centering
\vspace{-5pt}
\renewcommand{\arraystretch}{1.1}
\caption{Results of different methods and prompting on the EndoVis18 dataset.}
\vspace{5pt}
\setlength{\tabcolsep}{10pt} % Adjust the horizontal space between columns
\begin{tabular}{|c|c|c|}
% \hline
\hline\noalign{
\vskip-1
\doublerulesep}
\textbf{\makecell[c]{\rule{0pt}{10pt}Method Category}} & \textbf{\makecell[c]{\rule{0pt}{10pt}Method}} & \textbf{\makecell[c]{\rule{0pt}{10pt}CIoU}} \\ \hline

\multirow{5}{*}{\makecell[c]{Task-specific \\ Model}} 
 & TernausNet \cite{ref_ternausnet} & 46.2 \\ \cline{2-3}
 & MF-TAPNet \cite{ref_jin2019incorporating} & 67.9 \\ \cline{2-3}
 & ISINet \cite{ref_gonzalez2020in} & 73.0 \\ \cline{2-3}
 & S3Net \cite{ref_s3net} & 76.2 \\ \cline{2-3}
 & MATIS Frame \cite{ref_matis} & 82.4 \\ \hline

\multirow{12}{*}{\makecell[c]{SAM-based \\ Model}} 
 & MaskTrack-RCNN \cite{ref_maskrcnn} + SAM & 78.5 \\ \cline{2-3}
 & Mask2Former \cite{ref_mask2former} + SAM & 78.7 \\ \cline{2-3}
 & TrackAnything \cite{ref_trackanything} (1 Point) & 38.4 \\ \cline{2-3}
 & TrackAnything \cite{ref_trackanything} (5 Points) & 60.9 \\ \cline{2-3}
 & PerSAM \cite{ref_persam} & 49.2 \\ \cline{2-3}
 & SurgicalSAM \cite{ref_Yue2024surgsam} & 80.3 \\ 
 \cline{2-3} 
 \noalign{\vskip\doublerulesep}
 \cline{2-3}
 & SAM2 \cite{ref_Yu2024SAM2} (1 Point) & 63.6 \\ \cline{2-3}
 & SAM2 \cite{ref_Yu2024SAM2} (5 Points) & 78.8 \\ \cline{2-3}
 & SAM2 \cite{ref_Yu2024SAM2} (Full) & 82.2 \\ \cline{2-3}
 & SurgSAM2 (Ours) (1 Point) & 72.6 \\ \cline{2-3}
& SurgSAM2 (Ours) (5 Points) & 82.1 \\ \cline{2-3}
 & \textbf{SurgSAM2 (Ours)} (Full) & \textbf{84.4} \\ \hline
\end{tabular}
% \vspace{5pt}
% \vspace{-6pt}
\label{tab:iou_comparison}
\end{table*}

Similar to the performance in other evaluation matrices, we can see SurgSAM2 can consistently deliver superior results than the vanilla SAM2 in Challenge IoU, whether it is provided with detailed prompts (Full Mask) or more sparse prompts (1 Point and 5 Points). We also find that with the challenging instance-level setting, our method with mask prompt can achieve competitive IoU results, compared with these task-specific methods. More importantly, SurgSAM2 manages to achieve significant improvements in FPS and memory efficiency, superior to all the other methods.

In real-world surgical environments, both performance and computational demands need to be carefully considered. Given promising segmentation precision, coupled with high efficiency, our SurgSAM2 shows its great potential to facilitate the applicability of AI models in surgical deployment. Its consistent high performance across different prompting levels makes it a practical choice for integration into medical imaging workflows, providing surgeons with reliable and real-time segmentation results under various conditions.

\begin{comment}

\end{comment}

\subsubsection{Qualitative Evaluation}
\begin{figure*}[!ht]
    \centering
    \vspace{-10pt}
    \includegraphics[width=1\linewidth]{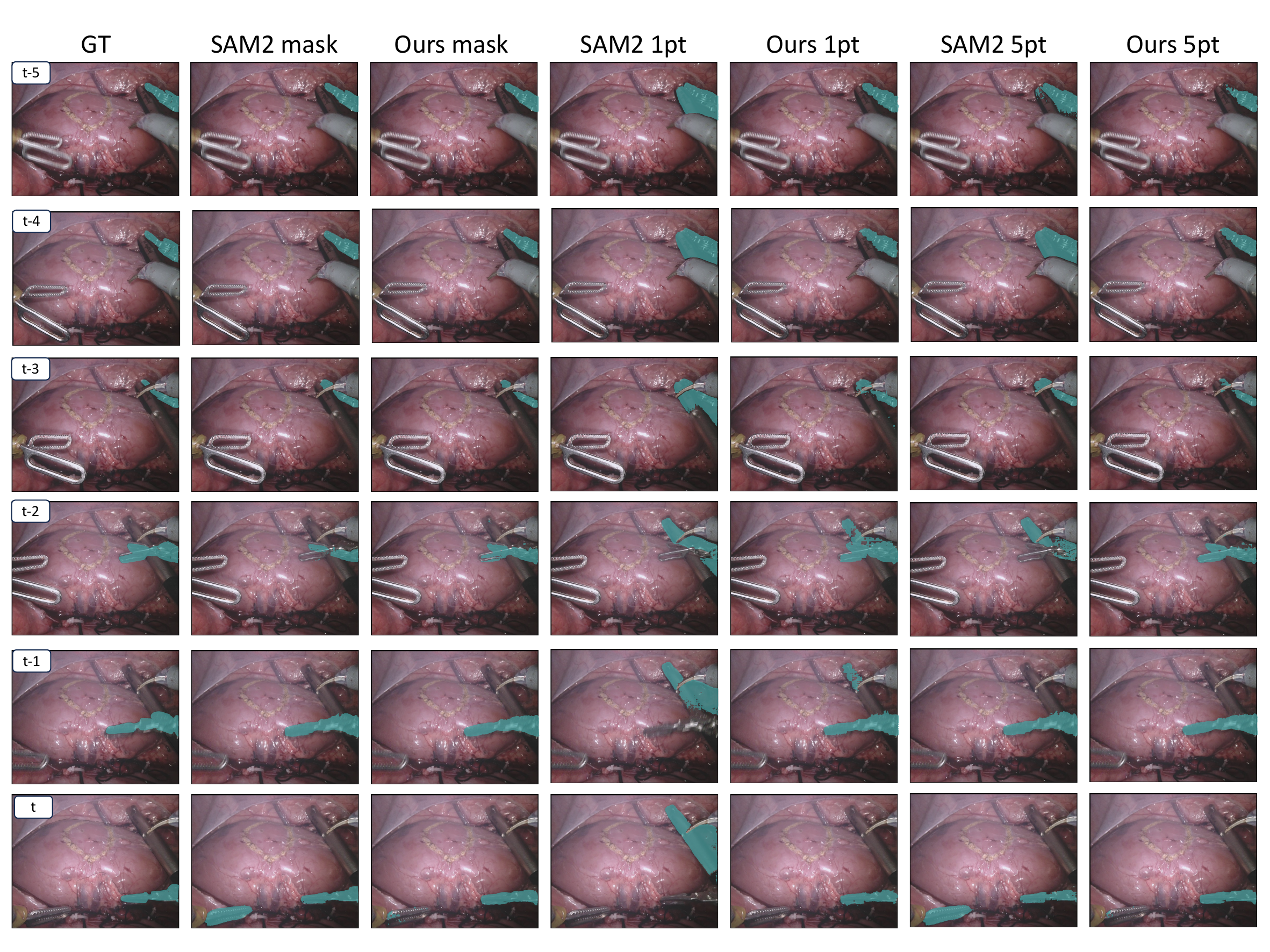}
    \vspace{-20pt}
    \caption{Visual comparison between SAM2 and SurgSAM2 on EndoVis18 dataset.}
    \label{fig:comparison}
    \vspace{-15pt}
\end{figure*}

Fig.~\ref{fig:comparison} presents a qualitative comparison between our SurgSAM2 and the vanilla SAM2 in full mask, one-point, and five-point prompt settings. The figure demonstrates that the vanilla SAM2 occasionally fails to segment the target object or identifies the incorrect object. On the contrary, our SurgSAM2 consistently produces accurate segmentation masks for the target instrument. Incorporating more accurate prompts, such as five-point or mask prompts, can further enhance segmentation accuracy.

\section{Conclusion and Future Work}

\vspace{-10pt}
In conclusion, our proposed model SurgSAM2 represents a significant advancement in the domain of surgical video segmentation. By integrating an EFP mechanism with the robust SAM2 framework, SurgSAM2 successfully addresses the challenges of real-time surgical video processing, enhancing both efficiency and accuracy. The ability of SurgSAM2 to selectively retain the most relevant frames based on cosine similarity has reduced memory usage while simultaneously improving the model’s segmentation performance across various tasks.

Our comprehensive evaluations on the EndoVis17 and EndoVis18 datasets demonstrate that SurgSAM2 consistently outperforms the vanilla SAM2 model, offering superior processing speed and reduced computational demands without compromising on accuracy. These results suggest that efficient memory management is crucial for advancing video segmentation in resource-constrained environments, particularly in the high-stakes context of surgical interventions.

Looking ahead, future research will focus on refining EFP strategies and experimenting with different memory bank sizes to identify the optimal configuration that maximizes both efficiency and segmentation accuracy. Additionally, we plan to expand the evaluation of SurgSAM2 across more diverse and complex datasets, further validating its robustness and applicability in various surgical contexts. By continuing to explore and integrate more sophisticated memory management techniques, we aim to push the boundaries of what is possible in real-time video analysis, not only within the medical field but also in broader applications that require rapid and accurate video segmentation.

\bibliographystyle{splncs04}
\bibliography{refs}
\end{document}